\documentclass[11pt]{article}

\usepackage[final]{acl}

\usepackage{times}
\usepackage{latexsym}

\usepackage[T1]{fontenc}

\usepackage[utf8]{inputenc}

\usepackage{microtype}

\usepackage{inconsolata}

\usepackage{graphicx}

\author{
  \textbf{Shangzhan Li\textsuperscript{1}},
  \textbf{Xinyu Yin\textsuperscript{2}},
  \textbf{Xuanyu Jin\textsuperscript{1}},
  \textbf{Ye He\textsuperscript{1}},
  \textbf{Yuxin Zhou\textsuperscript{1}},
\\
  \textbf{Yuxuan Li\textsuperscript{3}},
  \textbf{Xu Han\textsuperscript{3}},
  \textbf{Wanxiang Che\textsuperscript{1,$\dagger$}},
  \textbf{Qi Shi\textsuperscript{3,$\dagger$}},
  \textbf{Ting Liu\textsuperscript{1}},
  \textbf{Maosong Sun\textsuperscript{3}}
\\
\\
  \textsuperscript{1}Harbin Institute of Technology, Harbin, China\\
  \textsuperscript{2}Xiamen University, Xiamen, China\\
  \textsuperscript{3}Tsinghua University, Beijing, China
}

\usepackage{multirow}
\usepackage{multicol}
\usepackage{xspace}
\usepackage{amssymb}
\usepackage{bbm}
\usepackage{colortbl}
\usepackage{makecell}
\usepackage{graphicx}
\usepackage{booktabs}
\usepackage{amsmath}
\usepackage{arydshln}
\usepackage{amssymb}
\usepackage{tikz}
\usepackage{pgfplots}
\pgfplotsset{compat=1.8}
\usepgfplotslibrary{statistics}
\usepackage{makecell}
\usepackage{subcaption}

\usepackage{inconsolata}
\usepackage{hyperref}
\usepackage{amsfonts}
\usepackage[most]{tcolorbox}
\usepackage{dashrule}
\usepackage{ragged2e} 

\usepackage{inconsolata}

\newcommand\ours{\textsc{AutoVecCoder}\xspace}
\newcommand\vecprompt{\textsc{VecPrompt}\xspace}
\newcommand\vecrl{\textsc{VecRL}\xspace}

\title{\ours: Teaching LLMs to Generate Explicitly Vectorized Code}
\begin{document}

\maketitle
\renewcommand{\thefootnote}{$\dagger$}
\footnotetext{Corresponding authors.}
\renewcommand{\thefootnote}{\arabic{footnote}}

\begin{abstract}
Vectorization via Single Instruction, Multiple Data (SIMD) architectures is a cornerstone of high-performance computing. To fully exploit hardware potential, developers often resort to explicit vectorization using intrinsics, as compiler-based auto-vectorization frequently yields suboptimal results due to conservative static analysis. While Large Language Models (LLMs) have demonstrated remarkable proficiency in general code generation, they struggle with explicit vectorization due to the scarcity of high-quality corpora and the strict semantic constraints of low-level hardware instructions. In this paper, we propose \ours, a novel framework designed to empower LLMs with the capability of automated explicit vectorization. \ours integrates two core components: \vecprompt, an automated data synthesis pipeline to inject domain-specific intrinsic knowledge; and \vecrl, a reinforcement learning framework that aligns code generation with execution efficiency. 
\ours-8B trained by this framework achieves state-of-the-art performance on the SSE and AVX subsets of SimdBench and, in some cases, generates implementations surpassing standard \texttt{-O3} optimizations, effectively overcoming the inherent bottlenecks of traditional automated vectorization.
\end{abstract}

\section{Introduction}

\begin{figure}[ht]
\vspace{-5mm}
    \centering
    \includegraphics[width=0.4 \textwidth]{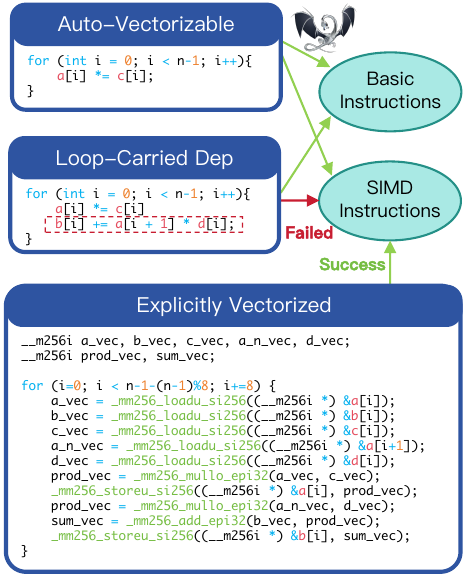}
    \caption{An example of explicit vectorization. From top to bottom, the figure shows: (1) code that can be automatically vectorized by the compiler; (2) code that cannot be automatically vectorized due to loop-carried data dependencies; and (3) explicitly vectorized code.}
    \label{fig:intro_code}
    \vspace{-3mm}
\end{figure}
Vectorization \citep{10.1145/3445814.3446692,10.1145/1133255.1133997,10.1145/1133981.1133997} is a fundamental programming paradigm for harnessing Data-Level Parallelism. By organizing data into vectors and leveraging Single Instruction, Multiple Data (SIMD) instruction sets, developers can process multiple data elements concurrently within a single clock cycle, achieving significant performance gains. From AVX \citep{intelintrinsics} in x86 architectures to SVE \citep{SVE-ARM} in ARM, vectorization has become an indispensable core technology for performance-sensitive applications, such as deep learning inference and scientific computing.

\begin{figure*}[t]
\vspace{-5mm}
    \centering
    \includegraphics[width=0.95 \textwidth]{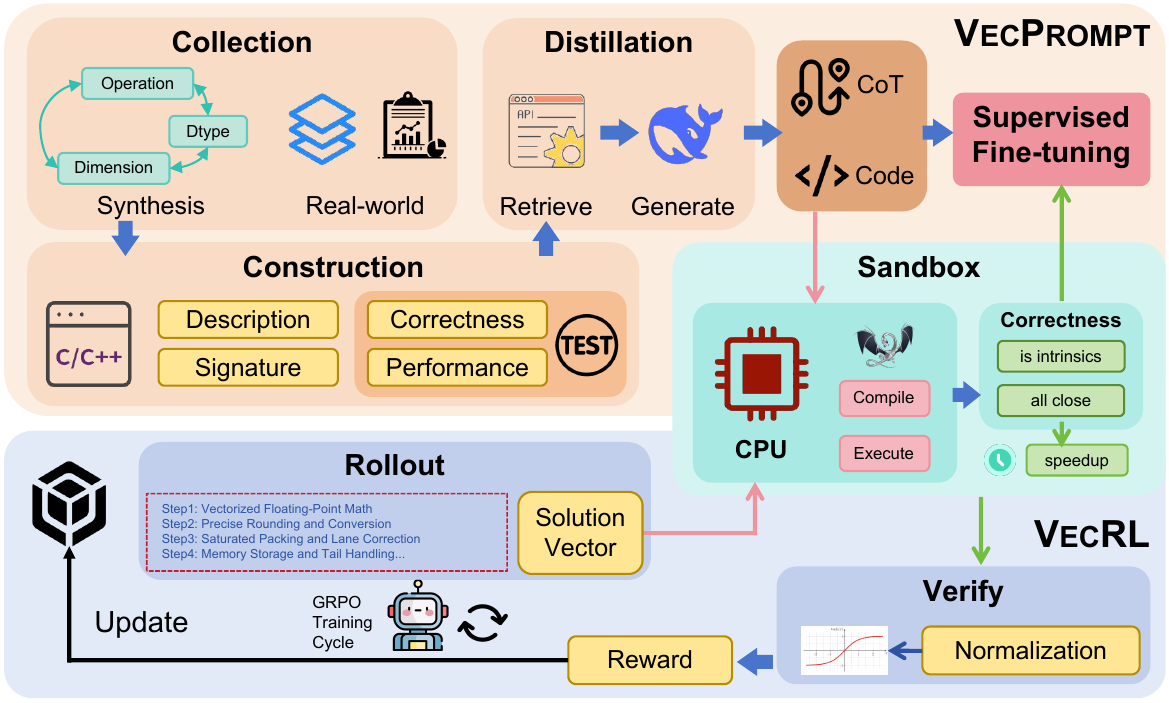}
    \caption{Overview of the \ours framework, which integrates knowledge-augmented data synthesis (\vecprompt) and performance-driven reinforcement learning (\vecrl) to enhance LLMs for explicit vectorization tasks.}
    \label{fig:main_pipeline}
\end{figure*}
Despite the immense parallel potential offered by modern hardware, generating efficient vectorized code remains a formidable challenge. Current development practices face a dichotomy between automation and performance. 
As shown in Figure~\ref{fig:intro_code}, implicit vectorization \citep{10.1145/2908080.2908111,10.5555/3454287.3455597} relies on compiler auto-vectorization; however, constrained by conservative static analysis, compilers often struggle to generate optimal instruction sequences. Conversely, explicit vectorization allows developers to directly control hardware via ISA-specific (Instruction Set Architecture) intrinsics, guaranteeing performance but introducing a steep learning curve, laborious coding processes, and limited portability. Consequently, automating the generation of high-performance, high-reliability explicitly vectorized code holds significant academic and industrial value.

Recent advancements in Large Language Models (LLMs) \citep{joel2025surveyllmbasedcodegeneration,zhang2024unifyingperspectivesnlpsoftware} have demonstrated remarkable capabilities in code generation. However, explicit vectorization presents unique challenges for LLMs due to the task's high logical density and strict adherence to low-level hardware constraints. Existing models frequently fail to generate code that is both semantically correct and optimized for specific instruction sets. For instance, SimdBench \citep{he2025simdbenchbenchmarkinglargelanguage} reveals substantial deficiencies in current models regarding high-performance vectorization. Furthermore, existing attempts \citep{10.1145/3696443.3708929} to build auto-vectorization workflows are hindered by an insufficient understanding of complex instruction mappings, making them ill-equipped to adapt to the rapid evolution of CPU architectures and SIMD instruction sets.

To tackle these challenges, we propose \ours, the first training framework deeply optimized for explicit vectorization tasks. This framework consists of two core components: 1) Automated Data Synthesis and Distillation Pipeline (\vecprompt): We construct a large-scale and reliable training corpus for vectorized programming by systematically constructing computational requirements and combining Retrieval-Augmented Generation (RAG) \citep{chen2025ultraragmodularautomatedtoolkit} mechanisms to inject high-quality domain knowledge of SIMD intrinsics. 2) Performance-Driven Reinforcement Learning Algorithm (\vecrl): By introducing a correctness-gated performance reward mechanism, we guide the model to optimize specifically towards generating "higher performance code."

Leveraging this framework, we develop \ours-8B model.\footnote{Model weights are available at \url{https://huggingface.co/LiShangZ/AutoVecCoder-8B}.}
Experimental results on the SSE and AVX subsets of SimdBench indicate that \ours, despite having only 8B parameters, achieves state-of-the-art performance compared with advanced closed-source models across a range of vectorization programming tasks.
Moreover, in some scenarios, it generates explicitly vectorized code that is superior to the results of compiler \texttt{-O3} optimization. This fully proves the effectiveness and advanced nature of our data pipeline and reinforcement learning algorithm.

In summary, the main contributions of this paper are as follows:
\begin{itemize}
    \item \vecprompt: We propose an automated data synthesis pipeline that utilizes RAG to inject SIMD intrinsic knowledge, constructing high-quality training data for vectorized programming.
    \item \vecrl: We introduce a GRPO-based reinforcement learning algorithm with a joint reward mechanism specifically designed to optimize both code correctness and execution performance.
    \item \ours-8B: We develop an 8B-parameter model that achieves SOTA performance compared with advanced closed-source models on the SSE and AVX subsets of SimdBench and, in specific scenarios, outperforms compiler \texttt{-O3} optimizations.
\end{itemize}

\section{Related Works}

The focus of LLM-based code generation has shifted from general functional correctness toward performance optimization. 
This section reviews recent advancements in LLM-driven vectorization and reinforcement learning for code efficiency, 
framing the context for \ours as a framework to achieve deep ISA-level alignment.

\subsection{LLM-based Vectorization Code Generation}
Recent research has increasingly explored the synergy between LLMs and vectorized programming to exploit the data parallelism of modern SIMD architectures (e.g., SSE, AVX, and AVX-512). 
Current approaches can be broadly categorized into implicit and explicit vectorization. Implicit vectorization focuses on program rewriting to improve the success rate of traditional compiler auto-vectorization. 
For instance, VecTrans \citep{zheng2025vectransenhancingcompilerautovectorization} utilizes an LLM-based agent to transform complex loop structures into compiler-friendly equivalent forms.
Explicit vectorization, conversely, involves the direct synthesis of code using SIMD intrinsics. 
LLM-Vectorizer \citep{10.1145/3696443.3708929} demonstrates the feasibility of this approach by translating scalar C programs into vectorized implementations, achieving speedups of up to $9.4\times$ on the TSVC \citep{6113845} benchmark. 
SimdBench \citep{he2025simdbenchbenchmarkinglargelanguage} provides a multi-architecture (x86, ARM, RISC-V) evaluation suite for assessing both functional correctness and execution performance. 
LLaMeSIMD \citep{vectorcamp-llamesimd-2025} and VecIntrinBench \citep{han2025vecintrinbenchbenchmarkingcrossarchitectureintrinsic} focus on the cross-architecture migration of SIMD intrinsics, with the latter specifically targeting the RISC-V vector extension. 
Building on this, IntrinTrans \citep{han2025intrintransllmbasedintrinsiccode} incorporates an execution-guided optimization workflow to refine intrinsic translation. 
However, these existing efforts primarily rely on zero-shot prompting or modular workflows, which lack a systematic training regime to inherently instill the specialized semantics of SIMD intrinsics into the model’s parameters.

\subsection{RL for Code Optimization}
As the proficiency of LLMs in code generation matures, the optimization objective has shifted from mere functional correctness to execution performance. 
One line of work treats performance tuning as a general reasoning task. For example, Afterburner \citep{du2025afterburnerreinforcementlearningfacilitates} employs the Group Relative Policy Optimization (GRPO) \citep{shao2024deepseekmathpushinglimitsmathematical} algorithm with execution speed as the reward signal to iteratively enhance code efficiency. 
 \citet{feng2025bettercorrectnessefficiencycode} proposed a two-stage alignment process using DPO \citep{rafailov2024directpreferenceoptimizationlanguage} and RLOO \citep{stiennon2020learning} to decouple the optimization of correctness and performance. 
This trend is mirrored by the emergence of performance-oriented benchmarks such as ECCO \citep{waghjale2024ecco}, EFFIBENCH \citep{huang2024effibench}, and Mercury \citep{du2024mercury}.
Another research direction focuses on domain-specific code optimization for complex hardware scenarios. 
KernelBench \citep{ouyang2025kernelbenchllmswriteefficient} and TritonBench \citep{li-etal-2025-tritonbench} evaluate the efficiency of operator-level implementations. 
Specialized frameworks like CUDA-L2 \citep{su2025cudal2surpassingcublasperformance} and AutoTriton \citep{li2025autotritonautomatictritonprogramming} utilize execution feedback within the GRPO framework to guide the generation of high-performance DSL code. 
Similarly, SUPERCODER \citep{wei2025supercoderassemblyprogramsuperoptimization} applies RL to assembly code optimization, and ChipSeek-R1 \citep{chen2025chipseekr1generatinghumansurpassingrtl} incorporates hardware PPA (Power, Performance, Area) metrics into a Verilog training pipeline.
These methods typically struggle with low-level DSL code, where strict architectural constraints, sparse optimization opportunities, and vast instruction combinations make effective performance-oriented code generation particularly challenging.

\section{Methodology}

Developing \ours consists of two primary stages: (1)\vecprompt, a knowledge-augmented distillation pipeline that synthesizes high-quality scalar-to-vector parallel corpora; 
and (2)\vecrl, a performance-driven reinforcement learning stage that aligns the model with efficiency using execution feedback.

\subsection{\vecprompt: Knowledge-Augmented Data Synthesis}
Explicit vectorization requires a precise mapping between scalar logic and architectural intrinsics. 
\vecprompt addresses the scarcity of such data by combining synthetic schemata with real-world code snippets, augmented by domain-specific knowledge retrieval.

\subsubsection{Seed Program Construction}
We first construct a diverse set of scalar C/C++ functions as source programs. 
To balance computational coverage and structural diversity, we employ a dual-source strategy.

\paragraph{Synthetic Schemata} We formalize vectorization requirements as a triplet $\mathcal{C} = \langle \mathcal{O}, \mathcal{T}, \mathcal{D} \rangle$, 
 where $\mathcal{O}$ denotes the operator type (e.g., GEMM, element-wise, reduction), $\mathcal{T}$ represents the data type (float, double, int, etc.), 
 and $\mathcal{D}$ specifies the input dimensions. 
 We further inject complex control flows (e.g., conditional branches) into these templates to simulate scenarios where traditional compiler auto-vectorization typically fails.
 
\paragraph{Real-world Collection} We harvest C/C++ snippets from established benchmarks like MBPP \citep{austin2021program} and XLCoST \citep{zhu2022xlcost}, normalizing them into a consistent functional format to ensure engineering compatibility.
Each seed function is annotated with a natural language description and a dedicated test suite for functional correctness and performance validation.

\subsubsection{Distillation via RAG}
To ensure the quality of distilled vectorized code, we incorporate an RAG mechanism to infuse the model with up-to-date hardware expertise. 
We construct a specialized knowledge base $\mathcal{K}$ from official SIMD intrinsic documentations. 
For each scalar function $f$, we perform dense semantic retrieval to obtain the top-$k$ relevant intrinsic definitions $\mathcal{K}_f^{(k)} = \{ d_1, \dots, d_k \}$ based on embedding similarity.
By providing $\mathcal{K}_f^{(k)}$ as context, the model generates vectorized code along with intermediate reasoning steps. 
This "knowledge-in-the-loop" approach significantly mitigates hallucinations regarding ISA-specific constraints. A case study to illustrate the role of RAG is shown in Appendix~\ref{appendix:case_study_rag}.

\subsubsection{Execution-Based Quality Control}
To guarantee the reliability of the training set, all generated candidates $\hat{f}$ undergo a rigorous filtering pipeline: 
(1) \textbf{Compilability}: The code must successfully compile for the target instruction set.
(2) \textbf{Functional Equivalence}: The output of $\hat{f}$ must match the scalar original across all test cases.
(3) \textbf{Complexity Constraint}: Candidates exceeding predefined length thresholds are discarded to avoid degenerate implementations.


The resulting high-fidelity dataset, $\mathcal{D}_{\text{SFT}} = \{(f, \mathcal{K}_f^{(k)}, \hat{f})\}$, is used for Supervised Fine-Tuning (SFT), establishing a robust baseline for instruction following and syntax correctness.

\subsection{\vecrl: Performance-Guided Reinforcement Learning}
While SFT establishes a foundational capability for instruction following and syntactical mapping, it often fails to capture the intricate performance landscapes of ISA execution. 
To bridge the gap between syntactically correct and computationally optimal code, we introduce \vecrl. 
This reinforcement learning stage shifts the optimization objective from static token-matching to dynamic execution efficiency, enabling the model to autonomously explore vectorization strategies that maximize performance gains on target architectures.

\subsubsection{Correctness-Constrained Reward Shaping}
\label{section:reward}
We model vectorization as a conditional policy optimization problem. 
A unique challenge in this domain is that performance feedback often exhibits a heavy-tailed distribution: minor code mutations may yield order-of-magnitude speedups. 
To stabilize training, we design a hierarchical reward function.
We first define the relative execution improvement $\Delta$:
\begin{equation}
\Delta = \frac{T_{\text{scalar}} - T_{\text{vector}}}{T_{\text{scalar}} + \epsilon}
\end{equation}
where $T_{\text{scalar}}$ and $T_{\text{vector}}$ denote the execution latency of the scalar and vectorized implementations, respectively.

To ensure stable policy convergence and distinguish between correctness and efficiency, our reward design follows three primary objectives: 
(1) \textbf{Strict Filtering}: Non-functional code is assigned zero reward to penalize syntax or logical errors.
(2) \textbf{Baseline Incentive}: A constant reward is provided for any functionally correct implementation to prevent the vanishing gradient problem.
(3) \textbf{Saturation-Aware Scaling}: Performance gains are rewarded within a bounded, Lipschitz-continuous range to prevent extreme outliers from dominating policy updates.


To satisfy these requirements, we formulate the total reward $R_{\text{total}}$ as:
\begin{equation}
    R_{\text{total}} = \mathbb{I}(\text{correct}) \cdot (\beta_{\text{base}} + \beta_{\text{perf}} \cdot \tanh(\alpha \cdot \Delta))
\end{equation}
where $\mathbb{I}(\cdot)$ is an indicator function for functional correctness. 
The coefficients $\beta_{\text{base}}$ and $\beta_{\text{perf}}$ balance the trade-off between basic instruction following and performance optimization, 
while the $\tanh$ mapping, scaled by a sensitivity factor $\alpha$, squashes the relative improvement $\Delta$ into a stable numerical manifold.

\subsubsection{Optimization with GRPO}
We employ Group Relative Policy Optimization (GRPO) to refine the model's policy. 
For each scalar function $f$, we generate a group of $G$ vectorized candidates $\{v_1, \dots, v_G\}$. 
The advantage $\hat{A}_i$ for each candidate is computed by normalizing its performance-aware reward $R_{\text{total}}$ (defined in Section~\ref{section:reward}) against the group statistics:
\begin{equation}
    \hat{A}_i = \frac{R_{\text{total}}(v_i) - \text{mean}({R_{\text{total}}})}{\text{std}({R_{\text{total}}}) + \epsilon}
\end{equation}
The model is optimized by maximizing the following objective, which incorporates the advantage and a KL divergence penalty to maintain training stability:
\begin{equation}
\begin{split}
\mathcal{L}_{\text{GRPO}} = & \mathbb{E}_{f, \{v_i\}} \Bigg[ \frac{1}{G} \sum_{i=1}^G \bigg( \frac{\pi_\theta(v_i \mid f)}{\pi_{\theta_{\text{old}}}(v_i \mid f)} \hat{A}_i \\
& - \eta D_{KL}(\pi_\theta \| \pi_{\text{ref}}) \bigg) \Bigg]
\end{split}
\end{equation}
This allows \ours to move beyond imitation of SFT data and discover novel ISA-level optimizations that transcend traditional compiler heuristics.

\section{Experiments}

In this section, we detail the experimental methodology designed to evaluate \ours. 
We first describe our two-stage data synthesis and training procedure, including the hyperparameters used for Supervised Fine-Tuning (SFT) and \vecrl. 
Subsequently, we outline our execution sandbox environment, which ensures resource isolation and precise performance measurement. 
Finally, we define the evaluation metrics and the set of state-of-the-art (SOTA) baselines used for comparative analysis.
\subsection{Experiment Setup}
\paragraph{Data Synthesis and SFT}
In the \vecprompt stage, we construct supervised training data for explicit vectorization through knowledge-augmented distillation. 
We employ the UltraRAG framework \citep{chen2025ultraragmodularautomatedtoolkit} as our retriever, utilizing a knowledge base built from official Intel SIMD intrinsics documentation \citep{intelintrinsics}. 
For each target scalar function, the retriever identifies the top-$5$ intrinsic snippets with the highest semantic relevance to serve as external expertise for the distillation process. 
We use DeepSeek-R1-250528 as the teacher model, resulting in a high-fidelity dataset of $7,685$ samples. Notably, this teacher model was frozen prior to the public release of SimdBench (July 2025), ensuring no data contamination between the distillation source and the evaluation benchmark. 
SFT is conducted using the LLaMA-Factory framework \citep{zheng-etal-2024-llamafactory} with a learning rate of $1\times10^{-5}$ over $3$ epochs, based on the Qwen3-8B \citep{yang2025qwen3technicalreport} model. Prompts during distillation and evaluation are shown in Appendix~\ref{appendix:prompts}. 
The resulting dataset exhibits broad coverage across multiple dimensions. In terms of computational semantics, it spans $6$ operator categories—arithmetic, math, logic/bitwise, reduction, type conversion, and comparison—with a near-uniform distribution drawn from a pool of $136$ unique operators. In terms of data representation, it covers $11$ element types including 32-bit/64-bit floating-point, signed and unsigned integers ranging from 8 to 64 bits, and string types, along with both 1D ($50.8$\%) and 2D ($49.2$\%) inputs. In terms of control-flow complexity, approximately $56$\% of the scalar reference programs contain $2$ or more loops (including nested structures over 2D inputs), and $32$\% include conditional branches, directly targeting the challenging scenarios where traditional compiler auto-vectorization typically fails.

\paragraph{Code Execution and RL}
During the \vecrl stage, we perform performance-driven policy optimization using the verl framework \citep{sheng2024hybridflow}.
The training set comprises $3,988$ samples, blending successfully distilled implementations with those that failed initial execution filtering to encourage robust exploration. 
We set the learning rate at $1\times10^{-6}$ with a batch size of $64$ for $5$ epochs. 
The reward function hyperparameters are configured as $\alpha = 3.0$, $\beta_{\text{base}} = 2.0$, and $\beta_{\text{perf}} = 1.0$. 
All RL experiments are conducted on a single node equipped with $8$$\times$NVIDIA A100 GPUs.
To ensure stable evaluation, we implement a lightweight execution sandbox using ZeroMQ (ZMQ) \citep{akgul2013zeromq} for task scheduling and resource management. 
Detailed implementation and benchmarking workflows are provided in Appendix~\ref{appendix:sandbox}.


\begin{table*}[ht!] \small
    \centering
    \resizebox{0.99\textwidth}{!}{
    \begin{tabular}{lcccccccc}
        \toprule
        \multirow{2}{*}{\textbf{Model}}& \multicolumn{4}{c}{\textbf{\textsc{AVX}}} & \multicolumn{4}{c}{\textbf{\textsc{SSE}}} \\
        \cmidrule(lr){2-5}
        \cmidrule(lr){6-9}
        & $\textbf{Corr}$ & $\textbf{fast}_\textbf{1}$ & $\textbf{P50}$ & $\textbf{P75}$ & $\textbf{Corr}$ & $\textbf{fast}_\textbf{1}$ & $\textbf{P50}$ & $\textbf{P75}$ \\
        \midrule
        Qwen3-Coder-480B-A35B& $55.15$ & $33.09$ & $0.67$ & $1.25$ & $45.59$ & $29.41$ & $-$ & $1.08$ \\
        Qwen3-Coder-Plus& $34.56$ & $19.85$ & $-$ & $0.89$ & $44.85$ & $25.00$ & $-$ & $1.00$ \\
        DeepSeek-R1-250528& $\underline{73.53}$ & $\underline{44.12}$ & $\underline{0.97}$ & $\underline{2.83}$ & $\underline{69.85}$ & $46.32$ & $\underline{0.99}$ & $1.95$ \\
        DeepSeek-V3-250324& $43.38$ & $24.26$ & $-$ & $1.00$ & $34.56$ & $22.79$ & $-$ & $0.97$ \\
        DeepSeek-V3.2-Thinking& $53.68$ & $30.88$ & $0.47$ & $1.30$ & $42.65$ & $27.94$ & $-$ & $1.01$ \\
        Gemini-2.5-Pro& $63.97$ & $39.71$ & $0.88$ & $\mathbf{2.84}$ & $61.76$ & $\underline{47.06}$ & $0.93$ & $\mathbf{2.35}$ \\
        GPT-5& $62.50$ & $36.76$ & $0.82$ & $1.18$ & $55.88$ & $33.82$ & $0.60$ & $1.15$ \\
        Grok4-Fast& $18.38$ & $9.56$ & $-$ & $-$ & $19.85$ & $11.03$ & $-$ & $-$ \\
        Claude-4-Sonnet-20250514& $58.09$ & $28.68$ & $0.72$ & $1.03$ & $66.91$ & $40.44$ & $0.93$ & $1.51$ \\
        \midrule
        Qwen3-8B (w/o Training) & $9.41$ & $2.79$ & $-$ & $-$ & $10.88$ & $5.00$ & $-$ & $-$ \\
        \ours-8B (w/o \vecrl) & $62.79$ & $35.59$ & $0.81$ & $1.85$ & $62.94$ & $43.53$ & $0.95$ & $1.70$ \\
        \ours-8B (Ours)& $\mathbf{76.76}$ & $\mathbf{47.35}$ & $\mathbf{0.99}$ & $2.74$ & $\mathbf{77.35}$ & $\mathbf{53.53}$ & $\mathbf{1.02}$ & $\underline{2.22}$ \\
        \bottomrule
    \end{tabular}
    }
    \caption{Main results on SimdBench comparing \ours-8B with various state-of-the-art LLMs across AVX and SSE instruction sets. Metrics include functional correctness (Corr), the proportion of correct samples with speedup $>1$ ($\text{fast}_1$), and the median (P50) and 75th percentile (P75) of speedup over correct samples. \ours-8B (w/o \vecrl) denotes the model after \vecprompt SFT only. Results for Qwen3-8B, \ours-8B (w/o \vecrl) are averaged over 5 runs. The best and second-best results are highlighted in \textbf{bold} and \underline{underlined}, respectively. ``$-$'' indicates insufficient correct samples for meaningful computation.}
    \label{tab:main_results}
\end{table*}
\subsection{Metrics and Baselines}
We evaluate execution efficiency using Google Benchmark library \citep{googlebench}. 
All generated code is compiled under the \texttt{-O3} optimization level to ensure a rigorous baseline. 
The primary metric, SpeedUp, is defined as the ratio of scalar execution time to vectorized execution time:
\begin{equation}
\text{SpeedUp} = \frac{T_{\text{scalar}}}{T_{\text{vector}}}
\end{equation}
where $T_{\text{scalar}}$ and $T_{\text{vector}}$ denote the execution time of the scalar and vectorized implementations, respectively, under identical input scales and hardware conditions. 
All benchmarks are executed on an Intel(R) Xeon(R) Platinum 8374C CPU @ 2.70GHz (x86\_64 architecture).
To characterize the overall proficiency of models in explicit vectorization, inspired by \citet{ouyang2025kernelbenchllmswriteefficient}, we use the $\text{fast}_\text{p}$ metric, which accounts for both functional correctness and performance gain:
\begin{equation}
\scalebox{0.92}{$
\text{fast}_\text{p} = \frac{1}{N} \sum_{i=1}^{N}\mathbbm{1}(\text{correct}_i \land \{\text{SpeedUp}_i > p\})
$}
\end{equation}
where $N$ is the total number of test cases, $\mathbbm{1}(\cdot)$ is the indicator function, and $p$ is a predefined performance threshold. 
This metric represents the percentage of samples that are both semantically correct and achieve a speedup exceeding $p$. 
To further characterize the speedup distribution, we report the median (P50) and 75th percentile (P75) of SpeedUp, computed exclusively over functionally correct samples.

To ensure a fair and contemporary comparison, we re-evaluated several SOTA models within our unified hardware environment. 
Following the SimdBench protocol, we evaluated: DeepSeek-V3-250324 \citep{deepseekai2025deepseekv3technicalreport}, DeepSeek-V3.2-Thinking \citep{liu2025deepseek}, DeepSeek-R1-250528 \citep{guo2025deepseek}, Qwen3-Coder-480B-A35B, Qwen3-Coder-Plus \citep{qwen3-coder-480b-a35b}, Gemini-2.5-Pro \citep{comanici2025gemini}, Grok4-Fast \citep{grok4-fast-xai-2025}, Claude-4-Sonnet \citep{anthropic-claude-sonnet-2025}, and GPT-5 \citep{openai-gpt5-2025}. 
Each model is tested under SSE (\texttt{-msse}, \texttt{-msse2}) and AVX (\texttt{-mavx}, \texttt{-mavx2}) instruction set configurations to assess their adaptability across different SIMD vector widths. 
We emphasize that all models are evaluated in a strictly zero-shot setting: neither \ours nor any baseline is provided with RAG-retrieved documentation during inference. The RAG-augmented knowledge injection is used exclusively during the \vecprompt data synthesis phase.



\section{Results}


\subsection{Main Results}
\label{section:main_results}
Table~\ref{tab:main_results} presents the comprehensive evaluation of \ours-8B on Simdbench, compared against a wide spectrum of advanced open-source and closed-source LLMs. 
Collectively, these results validate the efficacy of our \vecprompt + \vecrl training paradigm, 
demonstrating that a specialized 8B model can achieve—and even exceed—the domain-specific proficiency of massive, 
general-purpose frontier models through targeted data distillation and performance-driven reinforcement learning.

\paragraph{Superiority in Correctness and Performance}
Our proposed model, \ours-8B, achieves the highest correctness, $\text{fast}_\text{1}$, and median speedup (P50) on both instruction sets, while maintaining highly competitive P75 performance.
Notably, at the critical performance threshold of $\text{fast}_\text{1}$ (representing implementations that are strictly faster than their scalar counterparts), 
\ours-8B significantly outperforms much larger models, including DeepSeek-R1, Gemini-2.5-Pro, Claude-4, and GPT-5. 
This suggests that for low-level explicit vectorization tasks, performance-aware training objectives and execution-in-the-loop feedback are more decisive factors than raw parameter scale.

\paragraph{Synergistic Optimization of Correctness and Efficiency}
A key observation is that while some closed-source models achieve higher peak speedups (e.g., P75), their correctness rates are substantially lower. 
In contrast, \ours-8B consistently generates vectorized implementations that surpass the default \texttt{-O3} compiler optimizations while maintaining a high rate of functional equivalence. 
This underscores the advantage of our framework in navigating the correctness–performance trade-off, ensuring that generated code is not only fast but also reliable for production use.


\begin{figure}[t]
    \vspace{-5mm}
    \centering
    \includegraphics[width=0.48 \textwidth]{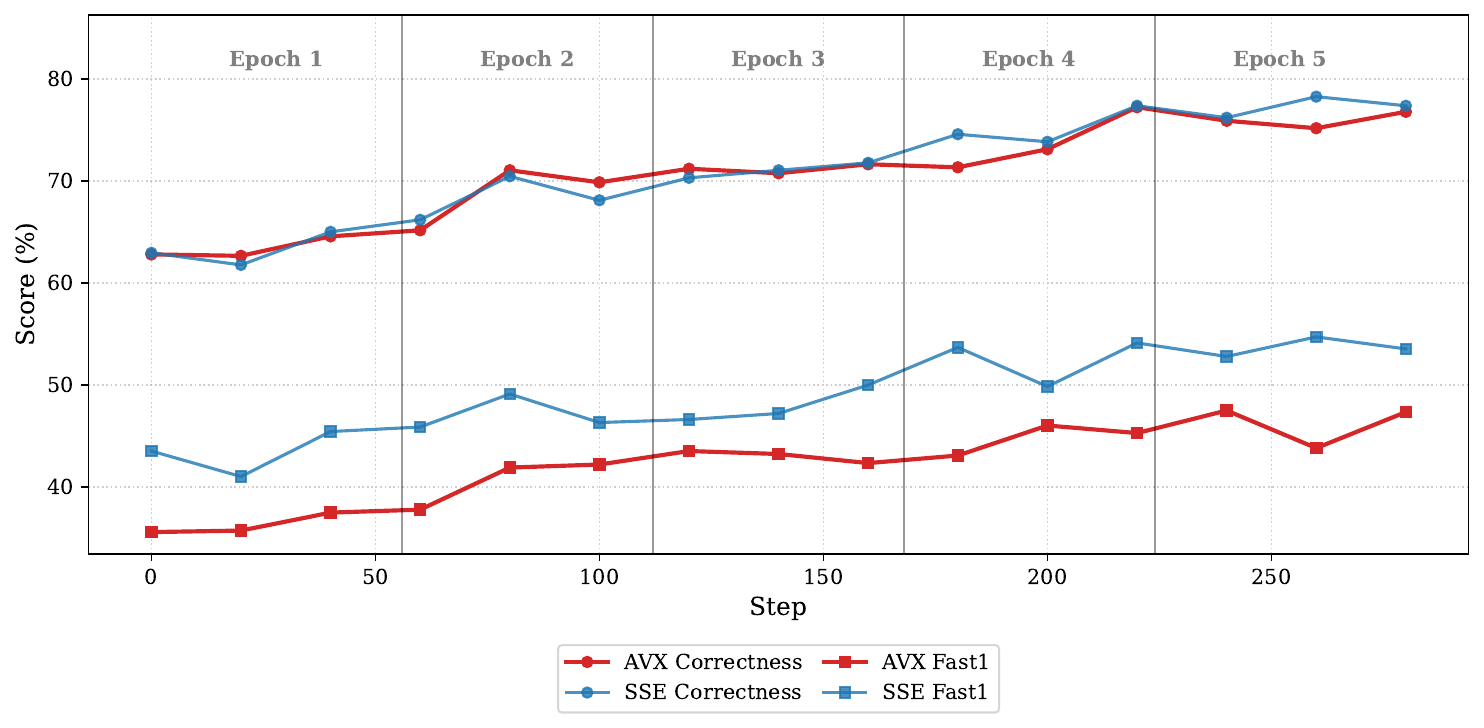}
    \caption{Performance evolution of \ours-8B during \vecrl, evaluated on the validation set every 20 optimization steps across 5 epochs. No smoothing is applied.}
    \label{fig:rl_iters}
\end{figure}

\subsection{Results Analysis}
\label{section:results_analysis}
\subsubsection{Performance Beyond \texttt{-O3}}
\label{section:performance_beyond}
We analyze cases in SimdBench where \ours-8B outperforms \texttt{-O3} auto-vectorization, identifying four recurring patterns where our model transcends the limitations of traditional compiler heuristics. The detailed case studies are shown in Appendix~\ref{appendix:case_study}.

\paragraph{Mask-based Control Flow}\label{case_study_1} Compilers often fail to vectorize loops with data-dependent branches due to conservative control-flow analysis. \ours-8B overcomes this by successfully transforming such logic into mask-based SIMD operations.
\paragraph{Handling Non-deterministic Iterations}\label{case_study_2} When loop structures prevent static inference of iteration behavior, compilers resort to scalar code. \ours-8B leverages SIMD-level condition handling to maintain parallelism without requiring rigid structural proofs.
\paragraph{Semantic Dependency Resolution}\label{case_study_3} Compilers are frequently inhibited by potential pointer aliasing or memory dependencies. \ours-8B utilizes its learned semantic intuition to safely apply vectorization where traditional analysis remains overly conservative.
\paragraph{Memory Access Restructuring}\label{case_study_4} Indirect or non-linear memory patterns are notoriously difficult for auto-vectorizers. \ours-8B demonstrates the ability to restructure these into SIMD-friendly access patterns, maximizing hardware throughput.

These patterns underscore the model's ability to exploit the semantic flexibility of SIMD intrinsics in scenarios where traditional compiler heuristics reach their theoretical limits.
\subsubsection{Training Dynamics of \vecrl}
To analyze the training dynamics of \vecrl, we track \ours's performance trajectory throughout the RL phase. 
Specifically, we evaluate the model every $20$ optimization steps across five epochs ($280$ steps total) using pass@1 correctness and $\text{fast}_\text{1}$ metrics on both AVX and SSE subsets. 
The resulting trends are illustrated in Figure~\ref{fig:rl_iters}.
\paragraph{Phase I: Functional Grounding}
During the first two epochs, we observe a significant surge in correctness for both instruction sets, with SSE and AVX correctness rapidly increasing from approximately $60$\% to $70$\%. 
However, the $\text{fast}_\text{1}$ metrics exhibit noticeable fluctuations without a consistent upward trend, 
suggesting that the policy is primarily guided by functional constraints, focusing on exploring the valid solution space of SIMD intrinsics.
\paragraph{Phase II: Performance Optimization}
A distinct shift occurs from the third epoch onward. While correctness continues to improve at a more gradual but stable rate, 
$\text{fast}_\text{1}$ for both SSE and AVX begins to show consistent and steady growth.
Notably, SSE $\text{fast}_\text{1}$ exhibits a particularly strong upward trend, 
indicating that the model effectively transitions from ``imitation of correctness'' to ``optimization for efficiency,'' 
prioritizing high-throughput SIMD implementations within the valid solution space.
\subsection{Analysis of Vectorization Depth}
\label{section:vectorization_depth}
Beyond the optimization patterns discussed above, we observe that the model learns vectorization strategies of varying depth, which warrants a nuanced examination.

\paragraph{Memory-Centric Vectorization}
For tasks that are inherently data-movement-bound (e.g., matrix row reordering, strided load/store), the model applies SIMD wide load/store operations as its primary strategy. Since these tasks involve no arithmetic computation, this constitutes the correct and complete vectorization approach, yielding measurable speedups through reduced instruction count and improved cache utilization.

\paragraph{Partial Vectorization}
In a smaller number of cases, the model uses SIMD instructions exclusively for batch data loading into temporary buffers, while the subsequent computation remains scalar. These implementations are technically valid uses of SIMD intrinsics, but the resulting speedup stems primarily from improved memory access patterns rather than computational parallelism.

\paragraph{Implications for Reward Design}
Neither pattern constitutes reward hacking, as the model does not exploit illegitimate shortcuts to inflate speedup metrics. However, these observations reveal that our current reward function optimizes for end-to-end execution speed without distinguishing the \emph{depth} of vectorization. Designing more fine-grained reward signals that assess the degree of computational parallelism remains a promising direction for future work.

\newcommand{\cmark}{\checkmark}

\begin{table*}[ht!] \small
    \centering
    \resizebox{0.9\textwidth}{!}{
    \begin{tabular}{cccccccccc}
        \toprule
        \multicolumn{2}{c}{\textbf{Training Stage}} & \multicolumn{4}{c}{\textbf{\textsc{AVX}}} & \multicolumn{4}{c}{\textbf{\textsc{SSE}}} \\
        \cmidrule(lr){1-2}
        \cmidrule(lr){3-6}
        \cmidrule(lr){7-10}
        \multirow{2}{*}{\vecprompt} & \multirow{2}{*}{\vecrl} & \multicolumn{2}{c}{\textbf{pass@1}} & \multicolumn{2}{c}{\textbf{pass@5}} & \multicolumn{2}{c}{\textbf{pass@1}} & \multicolumn{2}{c}{\textbf{pass@5}}  \\
        \cmidrule(lr){3-4}
        \cmidrule(lr){5-6}
        \cmidrule(lr){7-8}
        \cmidrule(lr){9-10}
        & & \textbf{Corr} & $\textbf{fast}_\textbf{1}$ & \textbf{Corr} & $\textbf{fast}_\textbf{1}$ & \textbf{Corr} & $\textbf{fast}_\textbf{1}$ & \textbf{Corr} & $\textbf{fast}_\textbf{1}$ \\

        \midrule
        &  & $9.41$ & $2.79$ & $20.59$ & $8.09$ & $10.88$ &  $5.00$  &  $26.47$ & $14.71$ \\
        & \cmark & $16.91$ & $7.35$ & $35.29$ & $15.44$ & $20.59$ & $9.56$ & $41.18$ & $22.79$ \\
        \cmark &  & $62.79$ & $35.59$ & $91.91$ & $61.76$ & $62.94$ & $43.53$ & $87.50$ & $70.59$ \\
        \cmark & \cmark & $\mathbf{76.76}$ & $\mathbf{47.35}$ & $\mathbf{93.38}$ & $\mathbf{69.85}$ & $\mathbf{77.35}$ & $\mathbf{53.53}$ & $\mathbf{94.85}$ & $\mathbf{75.74}$ \\
        \bottomrule
    \end{tabular}
    }
    \caption{Ablation study on the impact of \vecprompt in the training pipeline.}
    \label{tab:ablation_vecprompt}
\end{table*}
\subsection{Ablation Studies}
\label{section:ablation}
\subsubsection{Effect of \vecprompt{:} Bridging the Semantic Gap}
To analyze the role of \vecprompt in the overall training pipeline, we compare the reinforcement learning behavior under two settings: 
(1) direct \vecrl training on the base model (w/o \vecprompt), and (2) the full \ours pipeline (\vecprompt SFT followed by \vecrl). 
We track the pass@1, pass@5, and $\text{fast}_\text{1}$ metrics on the AVX and SSE instruction sets, as summarized in Table~\ref{tab:ablation_vecprompt}.
Experimental results demonstrate that \vecprompt provides a crucial initialization and stabilization effect for the reinforcement learning stage. 
This performance gap directly impacts the efficiency of \vecrl through two mechanisms:


\paragraph{Reward Sparsity} Higher initial correctness significantly increases the density of non-zero rewards during the rollout process. 
This allows the performance-driven signals to influence a larger proportion of training samples, leading to a more stable and faster convergence of the policy.

\paragraph{Search Space Constrainment} \vecprompt constrains the policy's exploration to a semantically reasonable subspace of SIMD intrinsics. 
By grounding the model in functionally correct implementations, the RL process can focus on differentiating and ranking implementations based on their execution efficiency, rather than struggling to learn basic syntax and SIMD semantics from scratch.

\subsubsection{Effect of Reward Function Design}

\begin{table}[ht!] \small
    \centering
    \resizebox{0.45\textwidth}{!}{
    \begin{tabular}{ccccc}
        \toprule
        \multirow{2}{*}{\textbf{Reward}} & \multicolumn{2}{c}{\textbf{\textsc{AVX}}} & \multicolumn{2}{c}{\textbf{\textsc{SSE}}} \\
        \cmidrule(lr){2-3}
        \cmidrule(lr){4-5}
        & \textbf{Corr} & $\textbf{fast}_\textbf{1}$ & \textbf{Corr} & $\textbf{fast}_\textbf{1}$ \\
        \midrule
        - & $62.79$ & $35.59$ & $62.94$ & $43.53$ \\
        NSR & $63.97$ & $36.03$ & $70.59$ & $46.32$ \\ 
        \vecrl & $\mathbf{69.85}$ & $\mathbf{41.18}$ & $\mathbf{72.06}$ & $\mathbf{47.79}$ \\
        \bottomrule
    \end{tabular}
    }
    \caption{Ablation study on the impact of \vecrl in the training pipeline.}
    \label{tab:ablation_vecrl}
\end{table}
To evaluate the impact of reward design on training stability and final performance, we compare our hierarchical reward against a Naive SpeedUp Reward (NSR), 
defined as 
\begin{equation}
NSR = \max\{1, \text{SpeedUp}\}.
\end{equation}
This baseline focuses solely on execution speed, falling back to a reward of 1 if the implementation is slower than the scalar version. 
Table~\ref{tab:ablation_vecrl} summarizes the performance of the model before RL and after one epoch of training under both reward settings.

Our proposed reward design consistently outperforms NSR across all metrics on both SSE and AVX instruction sets as shown in Table~\ref{tab:ablation_vecrl}, 
indicating that utilizing raw execution speed as the sole optimization signal is insufficient for generating reliable, high-quality vectorized code.
Detailed tracking of the optimization trajectory (shown in Figure~\ref{fig:ablation_vecrl}) reveals that NSR triggers a "cold-start" exploitation behavior. 
In the early stages of training, the model achieves a transient surge in performance by rapidly magnifying a narrow set of high-risk vectorized patterns that yield high immediate rewards but lack functional robustness. 
This speculative optimization leads to subsequent policy degradation; as training progresses, the model adopts increasingly fragile and over-specialized implementations that fail to generalize across diverse kernels. 
This results in a non-monotonic performance trend where initial gains are followed by a steady decline in both correctness and stability.

\section{Conclusion}


We presented \ours, a framework that empowers LLMs to generate high-performance vectorized code. 
By integrating knowledge-augmented distillation (\vecprompt) with performance-driven reinforcement learning (\vecrl), \ours bridges the gap between high-level code semantics and low-level hardware efficiency. 
Our evaluations show that \ours-8B maintains high functional correctness while uncovering optimization strategies beyond traditional compiler heuristics like \texttt{-O3}. 
This work demonstrates the efficacy of combining domain-specific knowledge with performance-driven feedback, offering a viable path for leveraging LLMs in specialized high-performance computing domains.

\section*{Limitations}
While \ours demonstrates high proficiency on x86 architectures (SSE/AVX), its generalizability to other SIMD instruction sets like ARM NEON or RISC-V Vector has not been extensively tested. 
Although our framework is conceptually architecture-agnostic, variations in instruction semantics and vector widths across platforms may pose challenges for knowledge retrieval and reward stability. 
Additionally, our current focus is on scalar C/C++ loops; extending the model to more complex, unstructured code or other high-performance DSLs remains future work. 

\section*{Ethics Statement}
This research utilizes publicly available benchmarks and synthesized code snippets, involving no human subjects or private data. 
We do not anticipate any significant ethical risks or negative social impacts. 
As with any automated programming tool, we recommend that generated code undergo standard functional verification and security auditing before deployment to ensure system stability and safety.

\section*{Acknowledgments}
We gratefully acknowledge the support of the National Natural Science Foundation of China (NSFC) via grant 62236004 and 62476073. 
This work was initiated and is supported by the AI9Stars Team.

\bibliography{custom,anthology-1,anthology-2}

\begin{thebibliography}{46}
\providecommand{\natexlab}[1]{#1}

\bibitem[{Akgul(2013)}]{akgul2013zeromq}
Faruk Akgul. 2013.
\newblock \emph{ZeroMQ}.
\newblock Packt Publishing.

\bibitem[{{Anthropic}(2025)}]{anthropic-claude-sonnet-2025}
{Anthropic}. 2025.
\newblock \href {https://www.anthropic.com/claude/sonnet} {Claude sonnet: Hybrid reasoning frontier model}.
\newblock \url{https://www.anthropic.com/claude/sonnet}.
\newblock Accessed: 2025-12-30.

\bibitem[{ARM(2025)}]{SVE-ARM}
ARM. 2025.
\newblock \href {https://developer.arm.com/documentation/102699/0100} {Sve optimization guide}.
\newblock Accessed: 2025-12-30.

\bibitem[{Austin et~al.(2021)Austin, Odena, Nye, Bosma, Michalewski, Dohan, Jiang, Cai, Terry, Le et~al.}]{austin2021program}
Jacob Austin, Augustus Odena, Maxwell Nye, Maarten Bosma, Henryk Michalewski, David Dohan, Ellen Jiang, Carrie Cai, Michael Terry, Quoc Le, and 1 others. 2021.
\newblock Program synthesis with large language models.
\newblock \emph{arXiv preprint arXiv:2108.07732}.

\bibitem[{Baghsorkhi et~al.(2016)Baghsorkhi, Vasudevan, and Wu}]{10.1145/2908080.2908111}
Sara~S. Baghsorkhi, Nalini Vasudevan, and Youfeng Wu. 2016.
\newblock \href {https://doi.org/10.1145/2908080.2908111} {Flexvec: auto-vectorization for irregular loops}.
\newblock In \emph{Proceedings of the 37th ACM SIGPLAN Conference on Programming Language Design and Implementation}, PLDI '16, page 697–710, New York, NY, USA. Association for Computing Machinery.

\bibitem[{Chen et~al.(2021)Chen, Mendis, Carbin, and Amarasinghe}]{10.1145/3445814.3446692}
Yishen Chen, Charith Mendis, Michael Carbin, and Saman Amarasinghe. 2021.
\newblock \href {https://doi.org/10.1145/3445814.3446692} {Vegen: a vectorizer generator for simd and beyond}.
\newblock In \emph{Proceedings of the 26th ACM International Conference on Architectural Support for Programming Languages and Operating Systems}, ASPLOS '21, page 902–914, New York, NY, USA. Association for Computing Machinery.

\bibitem[{Chen et~al.(2025{\natexlab{a}})Chen, Guo, Mei, Li, Chen, Li, Wang, Tang, Wang, Wu, Yan, Liu, Yu, Liu, and Sun}]{chen2025ultraragmodularautomatedtoolkit}
Yuxuan Chen, Dewen Guo, Sen Mei, Xinze Li, Hao Chen, Yishan Li, Yixuan Wang, Chaoyue Tang, Ruobing Wang, Dingjun Wu, Yukun Yan, Zhenghao Liu, Shi Yu, Zhiyuan Liu, and Maosong Sun. 2025{\natexlab{a}}.
\newblock \href {https://arxiv.org/abs/2504.08761} {Ultrarag: A modular and automated toolkit for adaptive retrieval-augmented generation}.
\newblock \emph{Preprint}, arXiv:2504.08761.

\bibitem[{Chen et~al.(2025{\natexlab{b}})Chen, Chang, Li, He, Chen, Li, Wang, Xu, Han, and Wang}]{chen2025chipseekr1generatinghumansurpassingrtl}
Zhirong Chen, Kaiyan Chang, Zhuolin Li, Xinyang He, Chujie Chen, Cangyuan Li, Mengdi Wang, Haobo Xu, Yinhe Han, and Ying Wang. 2025{\natexlab{b}}.
\newblock \href {https://arxiv.org/abs/2507.04736} {Chipseek-r1: Generating human-surpassing rtl with llm via hierarchical reward-driven reinforcement learning}.
\newblock \emph{Preprint}, arXiv:2507.04736.

\bibitem[{Comanici et~al.(2025)Comanici, Bieber, Schaekermann, Pasupat, Sachdeva, Dhillon, Blistein, Ram, Zhang, Rosen et~al.}]{comanici2025gemini}
Gheorghe Comanici, Eric Bieber, Mike Schaekermann, Ice Pasupat, Noveen Sachdeva, Inderjit Dhillon, Marcel Blistein, Ori Ram, Dan Zhang, Evan Rosen, and 1 others. 2025.
\newblock Gemini 2.5: Pushing the frontier with advanced reasoning, multimodality, long context, and next generation agentic capabilities.
\newblock \emph{arXiv preprint arXiv:2507.06261}.

\bibitem[{DeepSeek-AI et~al.(2025)DeepSeek-AI, Liu, Feng, Xue, Wang, Wu, Lu, Zhao, Deng, Zhang, Ruan, Dai, Guo, Yang, Chen, Ji, Li, Lin, Dai, Luo, Hao, Chen, Li, Zhang, Bao, Xu, Wang, Zhang, Ding, Xin, Gao, Li, Qu, Cai, Liang, Guo, Ni, Li, Wang, Chen, Chen, Yuan, Qiu, Li, Song, Dong, Hu, Gao, Guan, Huang, Yu, Wang, Zhang, Xu, Xia, Zhao, Wang, Zhang, Li, Wang, Zhang, Zhang, Tang, Li, Tian, Huang, Wang, Zhang, Wang, Zhu, Chen, Du, Chen, Jin, Ge, Zhang, Pan, Wang, Xu, Zhang, Chen, Li, Lu, Zhou, Chen, Wu, Ye, Ye, Ma, Wang, Zhou, Yu, Zhou, Pan, Wang, Yun, Pei, Sun, Xiao, Zeng, Zhao, An, Liu, Liang, Gao, Yu, Zhang, Li, Jin, Wang, Bi, Liu, Wang, Shen, Chen, Zhang, Chen, Nie, Sun, Wang, Cheng, Liu, Xie, Liu, Yu, Song, Shan, Zhou, Yang, Li, Su, Lin, Li, Wang, Wei, Zhu, Zhang, Xu, Xu, Huang, Li, Zhao, Sun, Li, Wang, Yu, Zheng, Zhang, Shi, Xiong, He, Tang, Piao, Wang, Tan, Ma, Liu, Guo, Wu, Ou, Zhu, Wang, Gong, Zou, He, Zha, Xiong, Ma, Yan, Luo, You, Liu, Zhou, Wu, Ren, Ren, Sha, Fu, Xu, Huang, Zhang, Xie, Zhang, Hao, Gou, Ma, Yan, Shao, Xu, Wu, Zhang, Li, Gu, Zhu, Liu, Li, Xie, Song, Gao, and Pan}]{deepseekai2025deepseekv3technicalreport}
DeepSeek-AI, Aixin Liu, Bei Feng, Bing Xue, Bingxuan Wang, Bochao Wu, Chengda Lu, Chenggang Zhao, Chengqi Deng, Chenyu Zhang, Chong Ruan, Damai Dai, Daya Guo, Dejian Yang, Deli Chen, Dongjie Ji, Erhang Li, Fangyun Lin, Fucong Dai, and 181 others. 2025.
\newblock \href {https://arxiv.org/abs/2412.19437} {Deepseek-v3 technical report}.
\newblock \emph{Preprint}, arXiv:2412.19437.

\bibitem[{Du et~al.(2024)Du, Luu, Ji, Liu, and Ng}]{du2024mercury}
Mingzhe Du, Anh~Tuan Luu, Bin Ji, Qian Liu, and See-Kiong Ng. 2024.
\newblock Mercury: A code efficiency benchmark for code large language models.
\newblock \emph{Advances in Neural Information Processing Systems}, 37:16601--16622.

\bibitem[{Du et~al.(2025)Du, Tuan, Liu, Qing, Huang, He, Liu, Ma, and kiong Ng}]{du2025afterburnerreinforcementlearningfacilitates}
Mingzhe Du, Luu~Anh Tuan, Yue Liu, Yuhao Qing, Dong Huang, Xinyi He, Qian Liu, Zejun Ma, and See kiong Ng. 2025.
\newblock \href {https://arxiv.org/abs/2505.23387} {Afterburner: Reinforcement learning facilitates self-improving code efficiency optimization}.
\newblock \emph{Preprint}, arXiv:2505.23387.

\bibitem[{Feng et~al.(2025)Feng, Xu, Xu, Hui, and Lin}]{feng2025bettercorrectnessefficiencycode}
Yunlong Feng, Yang Xu, Xiao Xu, Binyuan Hui, and Junyang Lin. 2025.
\newblock \href {https://arxiv.org/abs/2508.20124} {Towards better correctness and efficiency in code generation}.
\newblock \emph{Preprint}, arXiv:2508.20124.

\bibitem[{Google(2014)}]{googlebench}
Google. 2014.
\newblock \href {https://github.com/google/benchmark} {A microbenchmark support library}.
\newblock Originally released in 2014; accessed 2025.

\bibitem[{Guo et~al.(2025)Guo, Yang, Zhang, Song, Zhang, Xu, Zhu, Ma, Wang, Bi et~al.}]{guo2025deepseek}
Daya Guo, Dejian Yang, Haowei Zhang, Junxiao Song, Ruoyu Zhang, Runxin Xu, Qihao Zhu, Shirong Ma, Peiyi Wang, Xiao Bi, and 1 others. 2025.
\newblock Deepseek-r1: Incentivizing reasoning capability in llms via reinforcement learning.
\newblock \emph{arXiv preprint arXiv:2501.12948}.

\bibitem[{Han et~al.(2025{\natexlab{a}})Han, Kang, Xing, and Wu}]{han2025vecintrinbenchbenchmarkingcrossarchitectureintrinsic}
Liutong Han, Chu Kang, Mingjie Xing, and Yanjun Wu. 2025{\natexlab{a}}.
\newblock \href {https://arxiv.org/abs/2511.18867} {Vecintrinbench: Benchmarking cross-architecture intrinsic code migration for risc-v vector}.
\newblock \emph{Preprint}, arXiv:2511.18867.

\bibitem[{Han et~al.(2025{\natexlab{b}})Han, Tan, Zhang, Wang, Kang, Xing, and Wu}]{han2025intrintransllmbasedintrinsiccode}
Liutong Han, Zhiyuan Tan, Hongbin Zhang, Pengcheng Wang, Chu Kang, Mingjie Xing, and Yanjun Wu. 2025{\natexlab{b}}.
\newblock \href {https://arxiv.org/abs/2510.10119} {Intrintrans: Llm-based intrinsic code translator for risc-v vector}.
\newblock \emph{Preprint}, arXiv:2510.10119.

\bibitem[{He et~al.(2025)He, Zhao, Huang, Fu, Yu, Huang, and Xie}]{he2025simdbenchbenchmarkinglargelanguage}
Yibo He, Shuoran Zhao, Jiaming Huang, Yingjie Fu, Hao Yu, Cunjian Huang, and Tao Xie. 2025.
\newblock \href {https://arxiv.org/abs/2507.15224} {Simdbench: Benchmarking large language models for simd-intrinsic code generation}.
\newblock \emph{Preprint}, arXiv:2507.15224.

\bibitem[{Huang et~al.(2024)Huang, Qing, Shang, Cui, and Zhang}]{huang2024effibench}
Dong Huang, Yuhao Qing, Weiyi Shang, Heming Cui, and Jie~M Zhang. 2024.
\newblock Effibench: Benchmarking the efficiency of automatically generated code.
\newblock \emph{Advances in Neural Information Processing Systems}, 37:11506--11544.

\bibitem[{Intel(2025)}]{intelintrinsics}
Intel. 2025.
\newblock \href {https://www.intel.com/content/www/us/en/docs/intrinsics-guide/index.html} {Intel® intrinsics guide}.
\newblock Accessed: 2025-12-30.

\bibitem[{Joel et~al.(2025)Joel, Wu, and Fard}]{joel2025surveyllmbasedcodegeneration}
Sathvik Joel, Jie~JW Wu, and Fatemeh~H. Fard. 2025.
\newblock \href {https://arxiv.org/abs/2410.03981} {A survey on llm-based code generation for low-resource and domain-specific programming languages}.
\newblock \emph{Preprint}, arXiv:2410.03981.

\bibitem[{Li et~al.(2025{\natexlab{a}})Li, Li, Gao, Shi, Li, Wang, Huang, WangHaojie, Wang, Han, Liu, and Sun}]{li-etal-2025-tritonbench}
Jianling Li, ShangZhan Li, Zhenye Gao, Qi~Shi, Yuxuan Li, Zefan Wang, Jiacheng Huang, WangHaojie WangHaojie, Jianrong Wang, Xu~Han, Zhiyuan Liu, and Maosong Sun. 2025{\natexlab{a}}.
\newblock \href {https://doi.org/10.18653/v1/2025.findings-acl.1183} {{T}riton{B}ench: Benchmarking large language model capabilities for generating triton operators}.
\newblock In \emph{Findings of the Association for Computational Linguistics: ACL 2025}, pages 23053--23066, Vienna, Austria. Association for Computational Linguistics.

\bibitem[{Li et~al.(2025{\natexlab{b}})Li, Wang, He, Li, Shi, Li, Hu, Che, Han, Liu, and Sun}]{li2025autotritonautomatictritonprogramming}
Shangzhan Li, Zefan Wang, Ye~He, Yuxuan Li, Qi~Shi, Jianling Li, Yonggang Hu, Wanxiang Che, Xu~Han, Zhiyuan Liu, and Maosong Sun. 2025{\natexlab{b}}.
\newblock \href {https://arxiv.org/abs/2507.05687} {Autotriton: Automatic triton programming with reinforcement learning in llms}.
\newblock \emph{Preprint}, arXiv:2507.05687.

\bibitem[{Liu et~al.(2025)Liu, Mei, Lin, Xue, Wang, Xu, Wu, Zhang, Lin, Dong et~al.}]{liu2025deepseek}
Aixin Liu, Aoxue Mei, Bangcai Lin, Bing Xue, Bingxuan Wang, Bingzheng Xu, Bochao Wu, Bowei Zhang, Chaofan Lin, Chen Dong, and 1 others. 2025.
\newblock Deepseek-v3. 2: Pushing the frontier of open large language models.
\newblock \emph{arXiv preprint arXiv:2512.02556}.

\bibitem[{Maleki et~al.(2011)Maleki, Gao, Garzarán, Wong, and Padua}]{6113845}
Saeed Maleki, Yaoqing Gao, María~J. Garzarán, Tommy Wong, and David~A. Padua. 2011.
\newblock \href {https://doi.org/10.1109/PACT.2011.68} {An evaluation of vectorizing compilers}.
\newblock In \emph{2011 International Conference on Parallel Architectures and Compilation Techniques}, pages 372--382.

\bibitem[{Mendis et~al.(2019)Mendis, Yang, Pu, Amarasinghe, and Carbin}]{10.5555/3454287.3455597}
Charith Mendis, Cambridge Yang, Yewen Pu, Saman Amarasinghe, and Michael Carbin. 2019.
\newblock \emph{Compiler auto-vectorization with imitation learning}.
\newblock Curran Associates Inc., Red Hook, NY, USA.

\bibitem[{Nuzman et~al.(2006{\natexlab{a}})Nuzman, Rosen, and Zaks}]{10.1145/1133255.1133997}
Dorit Nuzman, Ira Rosen, and Ayal Zaks. 2006{\natexlab{a}}.
\newblock \href {https://doi.org/10.1145/1133255.1133997} {Auto-vectorization of interleaved data for simd}.
\newblock \emph{SIGPLAN Not.}, 41(6):132–143.

\bibitem[{Nuzman et~al.(2006{\natexlab{b}})Nuzman, Rosen, and Zaks}]{10.1145/1133981.1133997}
Dorit Nuzman, Ira Rosen, and Ayal Zaks. 2006{\natexlab{b}}.
\newblock \href {https://doi.org/10.1145/1133981.1133997} {Auto-vectorization of interleaved data for simd}.
\newblock In \emph{Proceedings of the 27th ACM SIGPLAN Conference on Programming Language Design and Implementation}, PLDI '06, page 132–143, New York, NY, USA. Association for Computing Machinery.

\bibitem[{{OpenAI}(2025)}]{openai-gpt5-2025}
{OpenAI}. 2025.
\newblock \href {https://openai.com/index/introducing-gpt-5/} {Introducing gpt-5}.
\newblock \url{https://openai.com/index/introducing-gpt-5/}.
\newblock Accessed: 2025-12-30.

\bibitem[{Ouyang et~al.(2025)Ouyang, Guo, Arora, Zhang, Hu, Ré, and Mirhoseini}]{ouyang2025kernelbenchllmswriteefficient}
Anne Ouyang, Simon Guo, Simran Arora, Alex~L. Zhang, William Hu, Christopher Ré, and Azalia Mirhoseini. 2025.
\newblock \href {https://arxiv.org/abs/2502.10517} {Kernelbench: Can llms write efficient gpu kernels?}
\newblock \emph{Preprint}, arXiv:2502.10517.

\bibitem[{{Qwen Team}(2025)}]{qwen3-coder-480b-a35b}
{Qwen Team}. 2025.
\newblock \href {https://qwenlm.github.io/blog/qwen3-coder/} {Qwen3-coder: Agentic coding in the world}.
\newblock Open source model release and technical blog.
\newblock Available from https://qwenlm.github.io/blog/qwen3-coder/.

\bibitem[{Rafailov et~al.(2024)Rafailov, Sharma, Mitchell, Ermon, Manning, and Finn}]{rafailov2024directpreferenceoptimizationlanguage}
Rafael Rafailov, Archit Sharma, Eric Mitchell, Stefano Ermon, Christopher~D. Manning, and Chelsea Finn. 2024.
\newblock \href {https://arxiv.org/abs/2305.18290} {Direct preference optimization: Your language model is secretly a reward model}.
\newblock \emph{Preprint}, arXiv:2305.18290.

\bibitem[{Shao et~al.(2024)Shao, Wang, Zhu, Xu, Song, Bi, Zhang, Zhang, Li, Wu, and Guo}]{shao2024deepseekmathpushinglimitsmathematical}
Zhihong Shao, Peiyi Wang, Qihao Zhu, Runxin Xu, Junxiao Song, Xiao Bi, Haowei Zhang, Mingchuan Zhang, Y.~K. Li, Y.~Wu, and Daya Guo. 2024.
\newblock \href {https://arxiv.org/abs/2402.03300} {Deepseekmath: Pushing the limits of mathematical reasoning in open language models}.
\newblock \emph{Preprint}, arXiv:2402.03300.

\bibitem[{Sheng et~al.(2024)Sheng, Zhang, Ye, Wu, Zhang, Zhang, Peng, Lin, and Wu}]{sheng2024hybridflow}
Guangming Sheng, Chi Zhang, Zilingfeng Ye, Xibin Wu, Wang Zhang, Ru~Zhang, Yanghua Peng, Haibin Lin, and Chuan Wu. 2024.
\newblock Hybridflow: A flexible and efficient rlhf framework.
\newblock \emph{arXiv preprint arXiv: 2409.19256}.

\bibitem[{Stiennon et~al.(2020)Stiennon, Ouyang, Wu, Ziegler, Lowe, Voss, Radford, Amodei, and Christiano}]{stiennon2020learning}
Nisan Stiennon, Long Ouyang, Jeffrey Wu, Daniel Ziegler, Ryan Lowe, Chelsea Voss, Alec Radford, Dario Amodei, and Paul~F Christiano. 2020.
\newblock Learning to summarize with human feedback.
\newblock \emph{Advances in neural information processing systems}, 33:3008--3021.

\bibitem[{Su et~al.(2025)Su, Sun, Li, Wang, Li, and Shum}]{su2025cudal2surpassingcublasperformance}
Songqiao Su, Xiaofei Sun, Xiaoya Li, Albert Wang, Jiwei Li, and Chris Shum. 2025.
\newblock \href {https://arxiv.org/abs/2512.02551} {Cuda-l2: Surpassing cublas performance for matrix multiplication through reinforcement learning}.
\newblock \emph{Preprint}, arXiv:2512.02551.

\bibitem[{Taneja et~al.(2025)Taneja, Laird, Yan, Musuvathi, and Lahiri}]{10.1145/3696443.3708929}
Jubi Taneja, Avery Laird, Cong Yan, Madan Musuvathi, and Shuvendu~K. Lahiri. 2025.
\newblock \href {https://doi.org/10.1145/3696443.3708929} {Llm-vectorizer: Llm-based verified loop vectorizer}.
\newblock In \emph{Proceedings of the 23rd ACM/IEEE International Symposium on Code Generation and Optimization}, CGO '25, page 137–149, New York, NY, USA. Association for Computing Machinery.

\bibitem[{{VectorCamp}(2025)}]{vectorcamp-llamesimd-2025}
{VectorCamp}. 2025.
\newblock \href {https://github.com/VectorCamp/LLaMeSIMD} {Llamesimd: The ultimate simd intrinsic \& function translation benchmarking suite}.
\newblock \url{https://github.com/VectorCamp/LLaMeSIMD}.
\newblock Accessed: 2025-12-30.

\bibitem[{Waghjale et~al.(2024)Waghjale, Veerendranath, Wang, and Fried}]{waghjale2024ecco}
Siddhant Waghjale, Vishruth Veerendranath, Zhiruo Wang, and Daniel Fried. 2024.
\newblock Ecco: Can we improve model-generated code efficiency without sacrificing functional correctness?
\newblock In \emph{Proceedings of the 2024 Conference on Empirical Methods in Natural Language Processing}, pages 15362--15376.

\bibitem[{Wei et~al.(2025)Wei, Suresh, Tan, Xu, Singh, Wang, and Aiken}]{wei2025supercoderassemblyprogramsuperoptimization}
Anjiang Wei, Tarun Suresh, Huanmi Tan, Yinglun Xu, Gagandeep Singh, Ke~Wang, and Alex Aiken. 2025.
\newblock \href {https://arxiv.org/abs/2505.11480} {Supercoder: Assembly program superoptimization with large language models}.
\newblock \emph{Preprint}, arXiv:2505.11480.

\bibitem[{{xAI}(2025)}]{grok4-fast-xai-2025}
{xAI}. 2025.
\newblock \href {https://x.ai/news/grok-4-fast} {Grok 4 fast}.
\newblock \url{https://x.ai/news/grok-4-fast}.
\newblock Accessed: 2025-12-30.

\bibitem[{Yang et~al.(2025)Yang, Li, Yang, Zhang, Hui, Zheng, Yu, Gao, Huang, Lv, Zheng, Liu, Zhou, Huang, Hu, Ge, Wei, Lin, Tang, Yang, Tu, Zhang, Yang, Yang, Zhou, Zhou, Lin, Dang, Bao, Yang, Yu, Deng, Li, Xue, Li, Zhang, Wang, Zhu, Men, Gao, Liu, Luo, Li, Tang, Yin, Ren, Wang, Zhang, Ren, Fan, Su, Zhang, Zhang, Wan, Liu, Wang, Cui, Zhang, Zhou, and Qiu}]{yang2025qwen3technicalreport}
An~Yang, Anfeng Li, Baosong Yang, Beichen Zhang, Binyuan Hui, Bo~Zheng, Bowen Yu, Chang Gao, Chengen Huang, Chenxu Lv, Chujie Zheng, Dayiheng Liu, Fan Zhou, Fei Huang, Feng Hu, Hao Ge, Haoran Wei, Huan Lin, Jialong Tang, and 41 others. 2025.
\newblock \href {https://arxiv.org/abs/2505.09388} {Qwen3 technical report}.
\newblock \emph{Preprint}, arXiv:2505.09388.

\bibitem[{Zhang et~al.(2024)Zhang, Chen, Liu, Liao, Gong, Yu, Li, and Wang}]{zhang2024unifyingperspectivesnlpsoftware}
Ziyin Zhang, Chaoyu Chen, Bingchang Liu, Cong Liao, Zi~Gong, Hang Yu, Jianguo Li, and Rui Wang. 2024.
\newblock \href {https://arxiv.org/abs/2311.07989} {Unifying the perspectives of nlp and software engineering: A survey on language models for code}.
\newblock \emph{Preprint}, arXiv:2311.07989.

\bibitem[{Zheng et~al.(2024)Zheng, Zhang, Zhang, Ye, and Luo}]{zheng-etal-2024-llamafactory}
Yaowei Zheng, Richong Zhang, Junhao Zhang, Yanhan Ye, and Zheyan Luo. 2024.
\newblock \href {https://doi.org/10.18653/v1/2024.acl-demos.38} {{L}lama{F}actory: Unified efficient fine-tuning of 100+ language models}.
\newblock In \emph{Proceedings of the 62nd Annual Meeting of the Association for Computational Linguistics (Volume 3: System Demonstrations)}, pages 400--410, Bangkok, Thailand. Association for Computational Linguistics.

\bibitem[{Zheng et~al.(2025)Zheng, Wu, Cheng, Li, Rocha, Liu, Wei, Zeng, Zhang, and Gao}]{zheng2025vectransenhancingcompilerautovectorization}
Zhongchun Zheng, Kan Wu, Long Cheng, Lu~Li, Rodrigo C.~O. Rocha, Tianyi Liu, Wei Wei, Jianjiang Zeng, Xianwei Zhang, and Yaoqing Gao. 2025.
\newblock \href {https://arxiv.org/abs/2503.19449} {Vectrans: Enhancing compiler auto-vectorization through llm-assisted code transformations}.
\newblock \emph{Preprint}, arXiv:2503.19449.

\bibitem[{Zhu et~al.(2022)Zhu, Jain, Suresh, Ravindran, Tipirneni, and Reddy}]{zhu2022xlcost}
Ming Zhu, Aneesh Jain, Karthik Suresh, Roshan Ravindran, Sindhu Tipirneni, and Chandan~K. Reddy. 2022.
\newblock \href {https://arxiv.org/abs/2206.08474} {Xlcost: A benchmark dataset for cross-lingual code intelligence}.
\newblock \emph{Preprint}, arXiv:2206.08474.

\end{thebibliography}

\appendix
\label{sec:appendix}

\section{The Design and Implementation of Execution Sandbox}
\label{appendix:sandbox}
\begin{figure}[ht]
\vspace{-5mm}
    \centering
    \includegraphics[width=0.45 \textwidth]{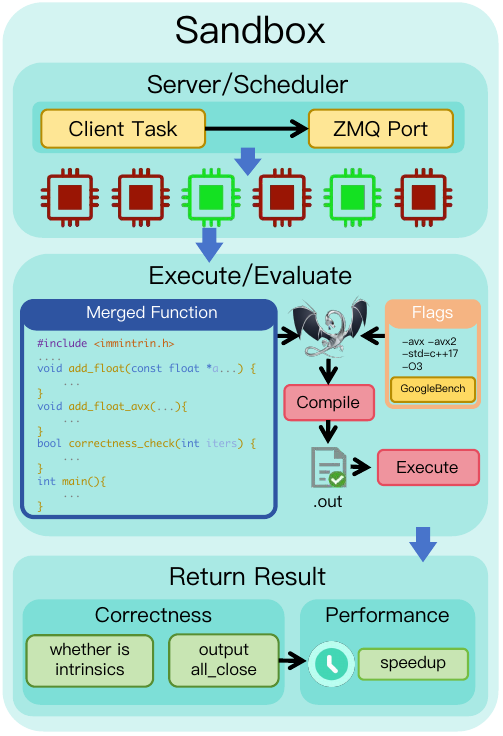}
    \caption{Architecture of sandbox for stable performance measurement.}
    \label{fig:sandbox}
    \vspace{-3mm}
\end{figure}
To provide stable and efficient feedback, we develop a high-concurrency execution sandbox with a tri-layer architecture: task scheduling, execution evaluation, and results retrieval.
Figure~\ref{fig:sandbox} illustrates the overall architecture of our execution sandbox.
The system utilizes ZeroMQ (ZMQ) for asynchronous communication and manages a pool of 64 dedicated physical CPU cores. 
To minimize measurement jitter and ensure isolation, each evaluation task is pinned to a specific core. 
This infrastructure supports the massive online evaluation requirements of the VecRL stage while maintaining high throughput and environment consistency.

The evaluation follows a rigorous "correctness-first" protocol. Generated candidates are compiled using clang++ and must first pass functional verification within a 20-second timeout. 
Only implementations that achieve semantic equivalence with the scalar baseline proceed to performance profiling. 
We utilize Google Benchmark with a 150-second timeout to measure execution latency. 
Each implementation is executed three times to mitigate measurement noise, using the average speedup as the final metric. 
This process ensures that the feedback used for filtering and reward calculation is both reliable and reproducible.

\section{The Performance of \vecrl and NSR in the First Epoch}
\begin{figure}[ht]
    \vspace{-5mm}
    \centering
    \includegraphics[width=0.45 \textwidth]{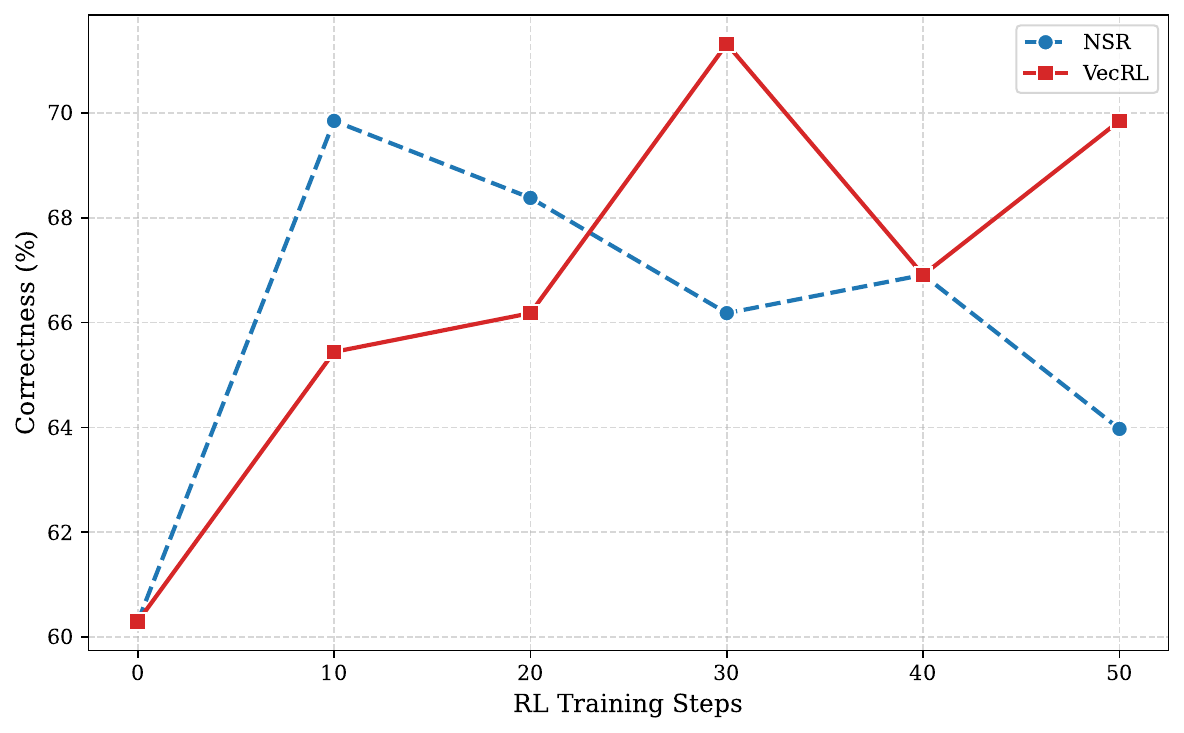}
    \caption{The Performance of \vecrl and NSR in the First Epoch.}
    \vspace{-3mm}
    \label{fig:ablation_vecrl}
\end{figure}

Figure~\ref{fig:ablation_vecrl} reveals a striking difference in optimization trajectories. 
In the early stages of training (approx. step 10), NSR leads to a temporary surge in both correctness and $\text{fast}_\text{1}$. 

\section{The Prompts of Distillation and Evaluation}
\label{appendix:prompts}
In this section, we provide the detailed prompts used for the distillation and evaluation stages. 
We select the translation task with scalar function implementation as input. All the evaluation prompts are same and listed in Figure~\ref{fig:appendix_prompt}.
\begin{figure*}
\centering
\begin{tcolorbox}[colframe=gray!80!black, colback=gray!10!white, title=Distillation and Evaluation Prompt]
\small
You are a code translator.\\
You will be given a C/C++ code snippet of a scalar implementation, and a function signature for the vectorized implementation with a description.\\
\hdashrule[0.5ex]{15cm}{0.1pt}{1mm}
\justifying
Your task is to translate the scalar code (source code) into vectorized code (target code) with \texttt{\{instruct set\}} intrinsics. \\
\hdashrule[0.5ex]{15cm}{0.1pt}{1mm}
\justifying \\
(\textbf{Only for Training})You may use the following simd-intrinsics API documentation to help you implement the function.\\
\texttt{\{retrieved\_simd\_api\_documents\}}\\
\hdashrule[0.5ex]{15cm}{0.1pt}{1mm}
\justifying \\
Return the SIMD code implementation within the markdown code block. \\
Do not include any explanations, comments, or text outside the code block. \\
\hdashrule[0.5ex]{15cm}{0.1pt}{1mm}
\justifying \\
Please provide the C/C++ code implementation for the following function using simd intrinsics: \\
Function Signature and Description:\\
\texttt{\{function signature\}}\\
Function Implementation: \\
\texttt{\{function implementation\}} \\
\end{tcolorbox}
\caption{Prompts used for distillation and evaluation.}
\label{fig:appendix_prompt}
\end{figure*}


\section{Case Studies of the Role of RAG}
\label{appendix:case_study_rag}
In this section, we present a case study (as shown in Figure~\ref{fig:case_study_rag}) that highlights the role of RAG in our framework.

\lstset{
    language=C,
    basicstyle=\ttfamily\scriptsize, 
    keywordstyle=\color{blue},
    commentstyle=\color{green!60!black},
    breaklines=true,
    linewidth=\textwidth,
    frame=single,
    numbers=left,
    numberstyle=\tiny\color{gray},
    numbersep=5pt,
    xleftmargin=15pt, 
    captionpos=t,
}

\begin{figure*}[t]
\centering
\begin{minipage}{0.95\textwidth}
\centering
\centering
\begin{lstlisting}[title={\small \textbf{SIMD-Optimized Implementation with RAG}:When using the RAG approach, the generated SIMD-optimized implementation includes the \_mm256\_min\_ps intrinsic function retrieved via RAG, which was not present in the version generated without RAG. After incorporating this intrinsic, the generated code becomes semantically correct. The retrieved content is provided in the comments of the following code block.}]
/*
signature: __m256 _mm256_min_ps (__m256 a, __m256 b)
description: Compare packed single-precision (32-bit) floating-point elements in 'a' and 'b', and store packed minimum values in ''dst''. [min_float_note]
operation: 
FOR j := 0 to 7
    i := j*32
    dst[i+31:i] := MIN(a[i+31:i], b[i+31:i])
ENDFOR
dst[MAX:256] := 0
*/
#include <immintrin.h>
#include <cfloat>
void argmin_relu_float_1D_simd(const float* A, float* result, size_t size) {
    if (size == 0) {
        *result = 0.0f;
        return;
    }
    __m256 min_vals = _mm256_set1_ps(FLT_MAX);
    __m256i min_indices = _mm256_setzero_si256();
    size_t i = 0;
    for (; i + 8 <= size; i += 8) {
        __m256 vA = _mm256_loadu_ps(A + i);
        __m256 v_relu = _mm256_max_ps(vA, _mm256_setzero_ps());
        __m256i v_index = _mm256_setr_epi32(i, i+1, i+2, i+3, i+4, i+5, i+6, i+7);
        __m256 mask = _mm256_cmp_ps(v_relu, min_vals, _CMP_LT_OS);
        min_vals = _mm256_min_ps(min_vals, v_relu);
        min_indices = _mm256_castps_si256(_mm256_blendv_ps(
            _mm256_castsi256_ps(min_indices),
            _mm256_castsi256_ps(v_index),
            mask
        ));
    }
    if (i < size) {
        size_t remainder = size - i;
        int mask_control[8] = {-1, -1, -1, -1, -1, -1, -1, -1};
        for (size_t j = remainder; j < 8; j++) {
            mask_control[j] = 0;
        }
        __m256i load_mask = _mm256_loadu_si256((const __m256i*)mask_control);
        __m256 vA = _mm256_maskload_ps(A + i, load_mask);
        __m256 v_relu = _mm256_max_ps(vA, _mm256_setzero_ps());
        v_relu = _mm256_blendv_ps(_mm256_set1_ps(FLT_MAX), v_relu, _mm256_castsi256_ps(load_mask));
        __m256i v_index = _mm256_setr_epi32(i, i+1, i+2, i+3, i+4, i+5, i+6, i+7);
        __m256 mask = _mm256_cmp_ps(v_relu, min_vals, _CMP_LT_OS);
        min_vals = _mm256_min_ps(min_vals, v_relu);
        min_indices = _mm256_castps_si256(_mm256_blendv_ps(
            _mm256_castsi256_ps(min_indices),
            _mm256_castsi256_ps(v_index),
            mask
        ));
    }
    __m128 min_vals_lo = _mm256_castps256_ps128(min_vals);
    __m128 min_vals_hi = _mm256_extractf128_ps(min_vals, 1);
    __m128 min1 = _mm_min_ps(min_vals_lo, min_vals_hi);
    __m128 min2 = _mm_min_ps(min1, _mm_permute_ps(min1, _MM_SHUFFLE(2,3,0,1)));
    __m128 min3 = _mm_min_ps(min2, _mm_permute_ps(min2, _MM_SHUFFLE(1,0,3,2)));
    float global_min_val = _mm_cvtss_f32(min3);
    __m256 v_global_min = _mm256_set1_ps(global_min_val);
    __m256 mask_eq = _mm256_cmp_ps(min_vals, v_global_min, _CMP_EQ_OQ);
    __m256i candidate_indices = _mm256_castps_si256(_mm256_blendv_ps(
        _mm256_castsi256_ps(_mm256_set1_epi32(size)),
        _mm256_castsi256_ps(min_indices),
        mask_eq
    ));
    __m128i candidate_lo = _mm256_castsi256_si128(candidate_indices);
    __m128i candidate_hi = _mm256_extracti128_si256(candidate_indices, 1);
    __m128i min_idx = _mm_min_epi32(candidate_lo, candidate_hi);
    __m128i min_idx1 = _mm_min_epi32(min_idx, _mm_shuffle_epi32(min_idx, _MM_SHUFFLE(2,3,0,1)));
    __m128i min_idx2 = _mm_min_epi32(min_idx1, _mm_shuffle_epi32(min_idx1, _MM_SHUFFLE(1,0,3,2)));
    int min_index = _mm_extract_epi32(min_idx2, 0);
    *result = (float)min_index;
}


\end{lstlisting}
\end{minipage}
\caption{Case study of the role of RAG.}
\label{fig:case_study_rag}
\end{figure*}

\section{Case Studies of \ours-8B Vectorization Patterns}
\label{appendix:case_study}
In this section, we present detailed case studies illustrating specific vectorization patterns learned by \ours-8B that enable it to surpass traditional compiler optimizations.
Figure~\ref{fig:case_study_mask_controlflow}, ~\ref{fig:case_study_handling_nondeterministic_iterations}, ~\ref{fig:case_study_semantic_dependency}, and ~\ref{fig:case_study_memory_access} 
showcase examples of mask-based control flow, handling non-deterministic iterations, semantic dependency resolution, and memory access restructuring described in Section~\ref{section:performance_beyond} respectively.

\lstset{
    language=C,
    basicstyle=\ttfamily\footnotesize, 
    keywordstyle=\color{blue},
    commentstyle=\color{green!60!black},
    breaklines=true,
    linewidth=0.95\textwidth,
    frame=single,
    numbers=left,
    numberstyle=\tiny\color{gray},
    numbersep=5pt,
    xleftmargin=15pt, 
    captionpos=t,
}


\begin{figure*}[t]
\centering
\begin{minipage}{0.9\textwidth}
\centering
\begin{lstlisting}[title={\small  \textbf{Conditional branches inside loops (Scalar implementation):} Perform element-wise three-valued logic operations sequentially over arrays.}]
void tensor_three_valued_logic(const int8_t* A, const int8_t* B, int8_t* C, size_t length) {
    for (size_t idx = 0; idx < length; idx++) {
        int a = A[idx], b = B[idx];
        if (a == 0 || b == 0) {
            C[idx] = 0;
        } else if (a > 0 && b > 0) {
            C[idx] = 1;
        } else if (a < 0 && b < 0) {
            C[idx] = -1;
        } else {
            C[idx] = 0;
        }
    }
}

\end{lstlisting}
\centering
\begin{lstlisting}[title={\small \textbf{Conditional branches inside loops (SIMD-optimized implementation)} : Use mask-based AVX SIMD operations to handle data-dependent conditional branches efficiently.}]
void tensor_three_valued_logic_simd(const int8_t* A, const int8_t* B, int8_t* C, size_t length) {
    size_t i = 0;
    const __m256i zero = _mm256_setzero_si256();
    const __m256i one = _mm256_set1_epi8(1);
    const __m256i minus_one = _mm256_set1_epi8(-1);

    for (; i + 31 < length; i += 32) {
        __m256i a = _mm256_loadu_si256((const __m256i*)(A + i));
        __m256i b = _mm256_loadu_si256((const __m256i*)(B + i));
        
        __m256i a_pos = _mm256_cmpgt_epi8(a, zero);
        __m256i b_pos = _mm256_cmpgt_epi8(b, zero);
        __m256i both_pos = _mm256_and_si256(a_pos, b_pos);
        
        __m256i a_neg = _mm256_cmpgt_epi8(zero, a);
        __m256i b_neg = _mm256_cmpgt_epi8(zero, b);
        __m256i both_neg = _mm256_and_si256(a_neg, b_neg);
        
        __m256i res = _mm256_and_si256(both_pos, one);
        res = _mm256_or_si256(res, _mm256_and_si256(both_neg, minus_one));
        
        _mm256_storeu_si256((__m256i*)(C + i), res);
    }

    for (; i < length; ++i) {
        if (A[i] > 0 && B[i] > 0) {
            C[i] = 1;
        } else if (A[i] < 0 && B[i] < 0) {
            C[i] = -1;
        } else {
            C[i] = 0;
        }
    }
}

\end{lstlisting}
\end{minipage}
\caption{Case study of mask-based control flow pattern learned by \ours-8B.}
\label{fig:case_study_mask_controlflow}
\end{figure*}

\begin{figure*}[t]
\centering
\begin{minipage}{0.9\textwidth}
\centering
\begin{lstlisting}[title={\small  \textbf{Uncertain loop structure and iteration behavior (Scalar implementation):} Check whether each row of a matrix is sorted according to specified directions executed sequentially.}]
bool matrix_rows_sorted_verify(const int* matrix, const bool* directions, size_t rows, size_t cols) {
    for (size_t i = 0; i < rows; i++) {
        bool ascending = directions[i];
        for (size_t j = 1; j < cols; j++) {
            int curr = matrix[i*cols + j];
            int prev = matrix[i*cols + (j-1)];
            if ((ascending && curr < prev) || (!ascending && curr > prev)) {
                return false;
            }
        }
    }
    return true;
}
\end{lstlisting}
\centering
\begin{lstlisting}[title={\small \textbf{Uncertain loop structure and iteration behavior (SIMD-optimized implementation)} : Use AVX SIMD intrinsics to perform batch-wise comparisons and condition handling without relying on loop-structure inference.}]
bool matrix_rows_sorted_verify_simd(const int* matrix, const bool* directions, size_t rows, size_t cols) {
    if (rows == 0 || cols <= 1) {
        return true;
    }
    for (size_t i = 0; i < rows; i++) {
        const int* row_ptr = matrix + i * cols;
        bool dir = directions[i];
        size_t num_pairs = cols - 1;
        size_t j = 0;
        if (num_pairs >= 8) {
            for (; j <= num_pairs - 8; j += 8) {
                __m256i v1 = _mm256_loadu_si256((const __m256i*)(row_ptr + j));
                __m256i v2 = _mm256_loadu_si256((const __m256i*)(row_ptr + j + 1));
                __m256i cmp;
                if (dir) {
                    cmp = _mm256_cmpgt_epi32(v1, v2);
                } else {
                    cmp = _mm256_cmpgt_epi32(v2, v1);
                }
                if (_mm256_movemask_epi8(cmp) != 0) {
                    return false;
                }
            }
        }
        for (; j < num_pairs; j++) {
            if (dir) {
                if (row_ptr[j] > row_ptr[j + 1]) {
                    return false;
                }
            } else {
                if (row_ptr[j] < row_ptr[j + 1]) {
                    return false;
                }
            }
        }
    }
    return true;
}
\end{lstlisting}
\end{minipage}
\caption{Case study of handling non-deterministic iterations learned by \ours-8B.}
\label{fig:case_study_handling_nondeterministic_iterations}
\end{figure*}

\begin{figure*}[t]
\centering
\begin{minipage}{0.9\textwidth}
\centering
\begin{lstlisting}[title={\small  \textbf{Conservative dependency analysis (Scalar implementation):} Perform a simple Caesar Cipher on a string sequentially.}]
std::string encrypt(const std::string & s){
    std::string out = '';
    int i;
    for (i=0;i<s.length();i++)
    {
        int w=((int)s[i]+4-(int)'a')%26+(int)'a';   
        out=out+(char)w;
    }
    return out;
}
\end{lstlisting}
\centering
\begin{lstlisting}[title={\small \textbf{Conservative dependency analysis (SIMD-optimized implementation):}  Use AVX SIMD intrinsics to apply character-wise shifts in parallel, bypassing conservative dependency assumptions.}]
std::string encrypt_simd(const std::string & s) {
    size_t len = s.length();
    if (len == 0) return '';
    std::string res(len, '\0');
    
    size_t i = 0;
    const size_t block_size = 32;
    if (len >= block_size) {
        for (; i <= len - block_size; i += block_size) {
            __m256i chunk = _mm256_loadu_si256(reinterpret_cast<const __m256i*>(s.data() + i));
            __m256i base = _mm256_set1_epi8('a');
            __m256i shifted = _mm256_sub_epi8(chunk, base);
            __m256i add_val = _mm256_set1_epi8(4);
            shifted = _mm256_add_epi8(shifted, add_val);
            __m256i cmp = _mm256_cmpgt_epi8(shifted, _mm256_set1_epi8(25));
            __m256i adjust = _mm256_and_si256(cmp, _mm256_set1_epi8(26));
            shifted = _mm256_sub_epi8(shifted, adjust);
            __m256i result = _mm256_add_epi8(shifted, base);
            _mm256_storeu_si256(reinterpret_cast<__m256i*>(res.data() + i), result);
        }
    }
    
    for (; i < len; ++i) {
        char c = s[i];
        c = (c - 'a' + 4) % 26 + 'a';
        res[i] = c;
    }
    
    return res;
}
\end{lstlisting}
\end{minipage}
\caption{Case study of semantic dependency resolution learned by \ours-8B.}
\label{fig:case_study_semantic_dependency}
\end{figure*}

\begin{figure*}[t]
\centering
\begin{minipage}{0.9\textwidth}
\begin{lstlisting}[title={\small  \textbf{Irregular memory access patterns (Scalar implementation) :} Compute the average of all elements in a matrix using non-linear memory accesses executed sequentially}]
double matrix_average(const double* mat, size_t m, size_t n) {
    double sum = 0.0;
    for (size_t j = 0; j < n; j++) {
        for (size_t i = 0; i < m; i++) {
            sum += mat[i * n + j];
        }
    }
    return sum / double(m * n);
}
\end{lstlisting}
\centering
\begin{lstlisting}[title={\small \textbf{Irregular memory access patterns (SIMD-optimized implementation)} : Restructure non-linear memory accesses into SIMD-friendly patterns using AVX intrinsics to improve parallelism.}]

double matrix_average_simd(const double* mat, size_t m, size_t n) {
    size_t total = m * n;
    if (total == 0) return 0.0;
    
    __m256d sum_vec = _mm256_setzero_pd();
    size_t i = 0;
    size_t vec_end = total - (total % 4);

    for (; i < vec_end; i += 4) {
        __m256d data = _mm256_loadu_pd(mat + i);
        sum_vec = _mm256_add_pd(sum_vec, data);
    }

    __m128d low = _mm256_extractf128_pd(sum_vec, 0);
    __m128d high = _mm256_extractf128_pd(sum_vec, 1);
    low = _mm_add_pd(low, high);
    __m128d sum_sse = _mm_hadd_pd(low, low);
    double sum = _mm_cvtsd_f64(sum_sse);

    for (; i < total; i++) {
        sum += mat[i];
    }

    return sum / (double)total;
}
\end{lstlisting}
\end{minipage}
\caption{Case study of memory access restructuring pattern learned by \ours-8B.}
\label{fig:case_study_memory_access}
\end{figure*}

\end{document}